# Emotion fusion for mental illness detection from social media: A survey

Tianlin Zhang [a], Kailai Yang [a], Shaoxiong Ji [b], Sophia Ananiadou [a,c,*]

[a] *National Centre for Text Mining, Department of Computer Science, The University of Manchester, Manchester, UK*
[b] *Department of Computer Science, Aalto University, Helsinki, Finland*
[c] *The Alan Turing Institute, London, UK*

A B S T R A C T

Mental illnesses are one of the most prevalent public health problems worldwide, which negatively influence people's lives and society's health. With the increasing popularity of social media, there has been a growing research interest in the early detection of mental illness by analysing user-generated posts on social media. According to the correlation between emotions and mental illness, leveraging and fusing emotion information has developed into a valuable research topic. In this article, we provide a comprehensive survey of approaches to mental illness detection in social media that incorporate emotion fusion. We begin by reviewing different fusion strategies, along with their advantages and disadvantages. Subsequently, we discuss the major challenges faced by researchers working in this area, including issues surrounding the availability and quality of datasets, the performance of algorithms and interpretability. We additionally suggest some potential directions for future research.

## 1. Introduction

Mental illness is a health condition that changes an individual's feelings, emotions, or behaviours. Currently, mental illnesses are one of the most prevalent public health concerns and remain a leading cause of disability and poor well-being worldwide [1]. The most common mental illnesses include depression, bipolar disorder, autism spectrum disorder (ASD), schizophrenia and other psychoses. According to a report released by the World Health Organisation (WHO),[1] one in eight people is living with one or more mental illnesses, resulting in a huge economic burden for governments. The COVID-19 pandemic has also exacerbated the problem [2]. Although effective treatment and prevention options exist, most patients with mental illnesses do not receive effective diagnosis and treatment due to ignorance of mental health assessment and the stigma surrounding mental illness [3]. Therefore, early detection of mental illness can prevent its progression into a serious state and can be beneficial to disease intervention.

With the exponential growth in the number of social media users over the past few years,[2] social media has become a primary source of information. Especially among young people, the use of social media platforms, such as Twitter, Reddit and Facebook, to share their thoughts and feelings, is becoming increasingly popular. Consequently, there has been a growing research interest in the detection of mental illness by analysing the user-generated textual content of social media posts. Indeed, a recent review on mental illness detection [4], reveals that the majority of research in this area (81%) is based on social media posts.

Recently, natural language processing (NLP) has begun to play an increasingly important role in social media data processing, and has been used to facilitate tasks such as sentiment analysis [5], rumor detection [6] and mental health screening [7]. Advances in NLP technologies and deep learning models explain the growing interest in exploring new methods for mental illness detection. For instance, deep learning-based methods such as Convolutional Neural Networks (CNN) [8], Recurrent Neural Networks (RNN) [9] are gradually replacing feature engineering-based methods, because deep learning frameworks enable models to automatically capture features without the need for time-consuming feature engineering. In addition, pre-trained Language Models (PLMs) such as BERT [10], RoBERTa [11] and MentalBERT [12], when trained on mental healthcare datasets, have achieved competitive performance on mental illness detection, which demonstrates their potential value.

Since emotions are an important part of human nature and can affect people's behaviours and mental states, they are widely studied in many research areas like behavioural science and psychology [13]. It has been found that there is a correlation between emotions and mental illnesses such as depression [14]. For example, some studies have shown that the severity of depressive symptoms is associated with an increasingly inverse relationship between positive and negative emotions [15]. The main reason is that people with depressive symptoms have difficulty regulating emotions, resulting in lower emotional complexity [16,17]. Therefore, from a psychological perspective, information about emotions is useful in diagnosing mental illness.

---

* Corresponding author at: National Centre for Text Mining, Department of Computer Science, The University of Manchester, Manchester, UK.
  *E-mail address:* sophia.ananiadou@manchester.ac.uk (S. Ananiadou).

[1] https://www.who.int/news-room/fact-sheets/detail/mental-disorders
[2] https://ourworldindata.org/rise-of-social-media

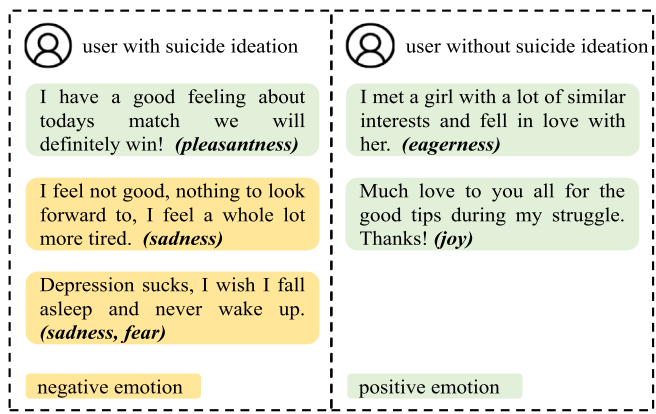

**Fig. 1.** Examples of tweeting history from two different users.

Psychologists have explored the impact of emotional factors from social media on mental health [18]. Several studies [19,20] have shown that the emotions expressed in posts written by subjects with mental illnesses differ from those written by control subjects. A study on identifying suicidal emotions in social media [21] also indicates that suicide-related posts are related to the user's psychological state, with negative emotions occurring significantly more frequently than positive emotions, as shown in Fig. 1. Such studies demonstrate the importance of leveraging emotions to support mental illness detection, and of combining or integrating them into medical applications.

The majority of existing surveys of mental illness detection review the application of NLP technologies to social media posts. For example, [22] presents a summary of data collection methods, classification models and evaluation results concerning mental health surveillance using social media, while [23] provides an analysis of computational methods for online mental state assessment. Other reviews have focused more specifically on deep learning-based methods [24,25] or the assessment of specific mental illnesses (e.g., suicide risk [26], depression [27]).

Recently, however, many studies (e.g. [28–30]) have focused on the role of emotion information in mental illness detection and have designed models that incorporate *emotion fusion*, which refers to the process of integrating or combining important emotion information with general textual information to obtain enhanced information for decision-making. To the best of our knowledge, this article constitutes the first survey of textual emotion information fusion for mental illness detection. We have carried out a comprehensive survey, with the aim of helping researchers to better understand the importance of emotion information in the detection of mental illnesses, and how the development of emotion fusion strategies can support this detection.

In this survey, we focus on strategies for fusing textual emotion information to support mental illness detection. State-of-the-art fusion strategies based on both conventional machine learning[3] and emerging deep learning approaches are studied. We also discuss the effects and challenges of different fusion strategies and highlight promising new research directions. The remainder of the paper is organised as follows: Section 2 introduces basic concepts of mental illness detection and recognising emotion information in social media. Section 3 presents and classifies different emotion fusion strategies to support mental illness detection. Section 4 provides some discussion about methods and fusion strategies. Section 5 describes the various challenges of emotion fusion and potential future directions of research. Finally, Section 6 concludes the paper.

## 2. Related work

### 2.1. Detection of mental illness in social media

From a clinical perspective, common data sources for detecting mental illness include clinical data, interviews, and surveys (like questionnaires [31] and rating scales [32]). However, these are costly and time-consuming to produce. Furthermore, although the majority of people with mental illness will not seek treatment [33], they will often express their moods or feelings through social media. Accordingly, the analysis of social media has become an important means to enhance understanding mental health problems. There are many potential applications for detecting mental illness from social media posts, including detecting whether someone has a mental disorder [30,34], predicting the severity of the disorder (mainly depression) [35,36], and exploring the causing factors of the disorder [37].

Fig. 2 depicts the general taxonomy of the different elements of mental illness detection in social media. The taxonomy consists of five categories, i.e., datasets, features, detection models and learning paradigms. Training datasets are obtained from various social media platforms like Twitter, Reddit or Facebook. Typically, some pre-processing of these datasets is undertaken, which may include tokenization, normalisation, removing stopwords and/or special words. The second category is features, which are divided into four main sets, i.e., linguistic features, domain knowledge features, statistical features and distributed representations. The first three features are usually extracted using NLP tools, while distributed representations have been widely used in deep learning-based models. Detection models can be classified into conventional machine learning-based methods and deep learning-based methods. The final category in the taxonomy, i.e., learning paradigms, consists of two main types, i.e., unsupervised and supervised learning, the latter of which can be further broken down into fully-supervised, semi-supervised and distantly supervised methods. The majority of mental illness detection models employ supervised learning methods, according to their ability to achieve superior performance when trained using high-quality datasets.

### 2.2. Emotion

Emotions are complex states of feeling caused by neurophysiological changes, which influence many aspects of human life. Charles Darwin hypothesised that emotion evolves along with natural selection and develops universal characteristics across races and cultures [38]. Accordingly, unified definitions of emotion have been proposed and are widely studied in psychology [39–41]. As emotion is often conveyed through natural language, the understanding of human emotions has attracted a large amount of research interest from the NLP community. Emotion recognition is a crucial aspect of many real-world applications. For example, dialogue systems for health care purposes require the machine to generate empathetic and emotionally coherent responses [42,43], which demands real-time emotion recognition of human utterances. Opinion mining in e-commerce [44,45] and social media analysis [46–48] also depend on the detection of emotion-related clues in customer reviews and social media posts.

A fundamental part of emotion recognition is the definition of appropriate emotion categories. Most work in this area defines emotions using two main types of models, i.e., categorical emotion models and dimensional emotion models. Categorical models divide emotions into a fixed set of discrete categories, while dimensional models define emotions as multi-dimensional continuous vectors, in which each dimension defines a corresponding aspect of the emotions. In terms of categorical definitions, [39] constitutes a pioneering work, which defines six basic emotion categories, i.e., happiness, surprise, sadness, anger, disgust and fear. [40] defined a more fine-grained set of eight primary emotion types, along with a wheel of emotions, in which each of these primary emotions is extended with multiple sub-types. In terms of dimensional definitions, the three most widely utilised dimensions are Valence, Arousal and Dominance (VAD), where Valence (V) reflects the pleasantness of a stimulus, Arousal (A) reflects the intensity of emotion provoked by a stimulus, and Dominance (D) reflects the degree of control exerted by a stimulus [49]. Dimensional categorisation represents each emotion as a point in the multi-dimensional space, which enables the annotators to carry out labelling using more fine-grained emotions rather than fixed emotion categories, and allows quantitative comparisons between emotions using vector operations such as inner product. As an example, we select three typical sentences from the EmoBank [50] dataset, where the sentences are labelled with three-dimensional VAD vectors and each score ranges from 1 to 5: the sentence '*lol Wonderful Simply Superb!*'(V: 4.6, A: 4.3, D:3.7) with the label *positive* is measurably closer in the representation space to the sentence '*I was feeling calm and private that night.*'(V: 3.1, A: 1.8, D: 3.1) with the label *neutral* than to the sentence '*Damn you!*' (V: 1.2, A: 4.2, D: 3.8) with the label *negative*. Moreover, Cambria et al. [41] proposed a psychologically-motivated hourglass model of emotions that represents affective states through emotion labels and four concomitant but independent dimensions (sensitivity, pleasantness, attention and aptitude), which can better represent the full range of emotions experienced by people.

---

[3] In this paper, we limit the conventional machine learning to be feature engineering-based methods that rely on manual feature extraction and selection, in contrast to representation learning-based deep learning methods.

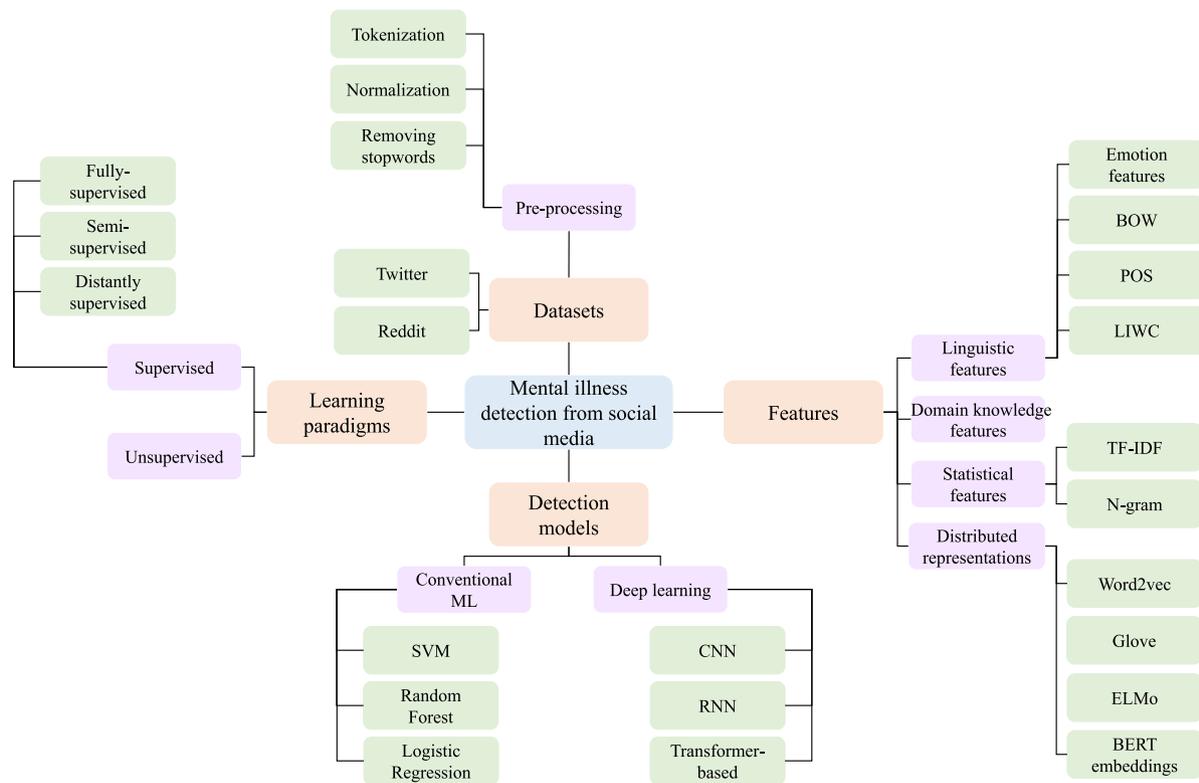

**Fig. 2.** Taxonomy of the different elements of detecting mental illness in social media.

Sentiment is also included within the scope of our study, due to its high correlation with emotion. Sentiment is a mental attitude or thought influenced by emotion. Sentiment analysis is a key NLP technology, which has been widely applied to social media, author identification, customer feedback, newswire and medical informatics [51–53]. According to different application scenarios, sentiment analysis systems may classify posts into positive/negative/neutral categories, or determine the strength of the sentiment.

## 3. Emotion fusion

Emotion fusion has become a hot topic in mental illness detection. In order to review research concerning the application of emotion fusion techniques to mental illness detection, we have performed a comprehensive literature search of articles published in a number of literature databases (i.e., IEEE Xplore, ACM Digital Library, Web of Science, PubMed, Scopus, and DBLP computer science bibliography). We firstly restricted the search to articles that concern the employment of NLP technologies to aid in the detection of mental illness in social media according to the search query [4]. We subsequently conducted a manual full-text review of the results of this restricted search, and retained articles mentioning the use of emotion fusion. This process resulted in the selection of 68 articles, which we further classify into feature engineering-based methods and deep learning-based methods in the next section.

Fig. 3 illustrates the distribution of the selected articles according to their date of publication. We can observe a recent upward trend in the number of relevant articles published, which demonstrates the increasingly important role of using emotion information in detecting mental illness. With the rise in the number of depressed users and suicides due to the COVID-19 pandemic [22,54], there has been a growing research interest in mental illness detection because of the importance and value of the task. In particular, the number of methods with emotion fusion is relatively high in 2021. Furthermore, due to the success of applying deep learning to NLP problems, the occurrence of deep learning-based approaches emotion fusion has increased since 2018.

### 3.1. Feature engineering-based methods

Methods based on feature engineering combine feature selection and extraction using NLP tools with machine learning methods. According to a recent review paper [4], a wide range of conventional machine learning-based models have been leveraged to detect mental disorders. Fig. 4 illustrates the process of the emotion fusion within feature engineering-based methods. Firstly, NLP tools are used to extract both emotion-based features and other features. Subsequently, machine learning models are applied to select and combine various features in order to carry out the final classification. In this section, we introduce various commonly used tools for the extraction of emotion features, and outline some of the conventional machine learning models that have been applied for mental illness detection. Table 1 provides an overview of the approaches to mental illness detection that use feature engineering in conjunction with emotion fusion.

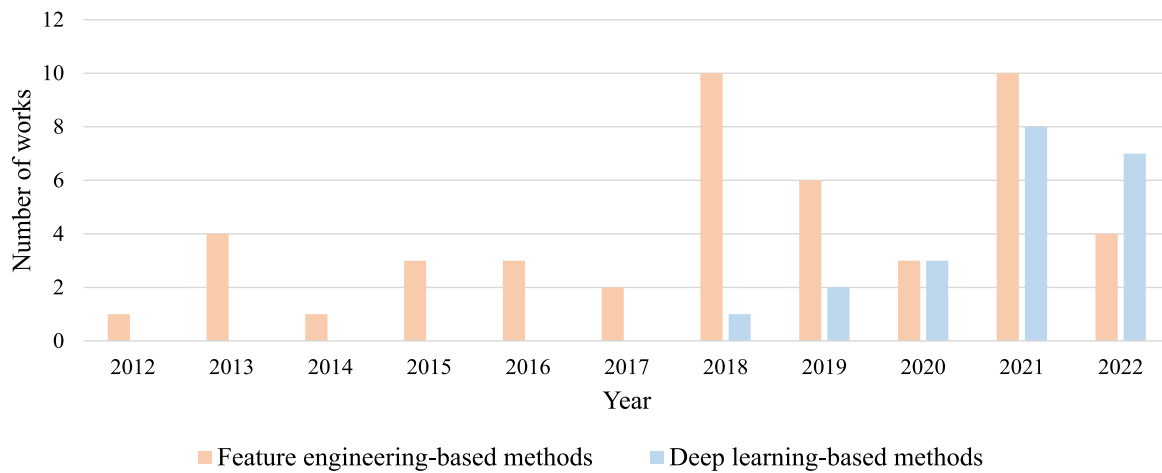

**Fig. 3.** Distribution of publications on emotion fusion for mental illness detection since 2012.

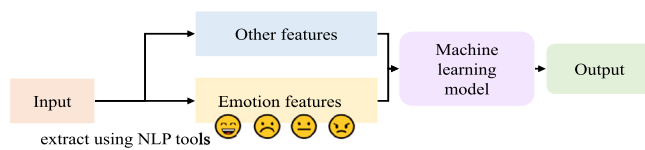

**Fig. 4.** Emotion fusion using feature engineering-based methods.

*3.1.1. Emotion feature extraction tools*

Below, we summarise the most commonly used emotion feature extraction tools and language resources that are used in feature engineering-based mental illness detection models.

**Linguistic Inquiry and Word Count (LIWC)**[4]: LIWC [55] is text analysis software that counts words belonging to psychological, linguistic and topical categories that demonstrate various affective, social and cognitive processes. The core of the program compares each word in a document to the list of LIWC dictionary words and calculates the proportion of words that belong to different categories. According to its convenience, many studies [56,57] have used LIWC to estimate emotional states expressed in the text.

**Affective norms for English words (ANEW)**: The ANEW lexicon [58] is a lexicon of words rated by human subjects according to a three-dimensional space of valence, arousal and dominance. In [59], the authors used the valence values of ANEW lexicon on a scale from 1 to 9 to facilitate the quantitative estimation of depression.

**NRC Emotion lexicon**: The NRC emotion lexicon [60] is a general emotion lexicon with 14,182 entries, which uses crowdsourcing to annotate words according to Plutchik's eight basic emotions (i.e., anger, fear, anticipation, trust, surprise, sadness, joy, and disgust) and two sentiments (positive and negative). In [61], the authors represented each individual language as normalised frequency distributions of these basic emotions using the NRC emotion lexicon.

**Valence Aware Dictionary for sEntiment Reasoning (VADER)**: VADER [62] is a sentiment lexicon and rule-based analysis tool for general sentiment analysis. The VADER sentiment lexicon contains scores (scale from −4 to 4) that denote both intensity and polarity. For instance, the valence of word 'good' is 1.9 indicating positive, and 'great' is 3.1, while 'horrible' is −2.5. The VADER sentiment analysis tool is a rule-based model combining lexical features and five general grammatical and syntactical rules for emotion intensity perception. For example, [63] employed VADER sentiment analysis to extract sentiment features.

**TextBlob**[5]: TextBlob is a Python sentiment analysis library that uses Natural Language ToolKit (NLTK). Given a sentence, TextBlob returns two types of output values, i.e., polarity and subjectivity. Polarity values lie between [−1, 1] according to the degree to which the sentiment is positive or negative. Subjectivity values lie between [0,1], indicating personal opinions and judgments. In [64], the authors imported the TextBlob library to calculate the sentiment scores with subject and polarity, and fed them into depression detection classifiers

**EMOTIVE**: EMOTIVE [65] is an ontology based system for recognising fine-grained emotions from social media. The emotion features used by EMOTIVE include eight basic emotions (anger, hear, confusion, disgust, happiness, sadness, shame, and surprise). An overall post's emotionality score is calculated by summing of all emotion scores. The study [66] found that the EMOTIVE feature set performs better than the LIWC feature set for bipolar and depression detection.

**SentiWordNet**: SentiWordNet [67] is a publicly available lexical resource for sentiment analysis and opinion mining applications. SentiWordNet assigns sentiment scores to all WordNet [68] synsets according to the notions of positivity, negativity, and objectivity. In [69], the authors calculated affective valence expressed in each entry using SentiWordNet.

**SentiStrength**[6]: SentiStrength [70] is a customised sentiment detection tool, which can estimate the strength of positive or negative sentiment within short, informal texts on social media. The tool reports sentiment strength on a scale from −5 (extremely negative) to 5 (extremely positive), which was used in [71].

**SenticNet**: SenticNet [72] is a publicly available semantic resource for concept-level sentiment analysis. SenticNet is automatically constructed by applying an ensemble of graph-mining and multi-dimensional scaling techniques on various affective commonsense knowledge resources, which results in 400,000 commonsense concepts. The latest version, SenticNet 7 [73], leverages commonsense-based neurosymbolic AI models. Several publications [74,75] used SenticNet to extract sentiment features for depression detection.

---
[4] https://www.liwc.app/
[5] https://textblob.readthedocs.io/
[6] http://sentistrength.wlv.ac.uk/

In additional to the tools and resources outlined above, a number of commercial sentiment analysis tools are also used for sentiment feature extraction. These include ParallelDots.[7] and MeaningCloud[8]

### 3.1.2. Conventional machine learning methods

In addition to emotion-based features, feature engineering-based approaches to mental illness detection employ a variety of other types of features that can easily be obtained by common NLP tools, including linguistic features and statistical features. Linguistic features include Bag-of-words (BOW) [76], Parts-of-Speech (POS) [77], and information obtained from the Linguistic Inquiry and Word Count (LIWC) software [56,78], while statistical features include term frequency-inverse document frequency (TF-IDF) [79] and n-gram [80].

Machine learning models are designed to train classifiers based on different combinations of features such as those introduced. As shown in Table 1, our survey has revealed that a variety of conventional machine learning methods have been used for mental illness detection, including Support Vector Machines (SVM) [81,82], Logistic Regression [61,83], Random Forest [63,76], Decision Trees [57,84] and Adaptive Boosting (Adaboost) [80,85]. Most of these machine learning models are based on a supervised learning framework, relying on high-quality labelled training data. Moreover, some works used semi-supervised learning techniques (i.e., models trained on small amounts of labelled data in conjunction with large amounts of unlabelled data) [71,86]. These semi-supervised techniques include the classical algorithms "Yet Another Two-Stage Idea" (YATSI) and "Learning with Local and Global Consistency" (LLGC) [71].

### 3.2. Deep learning-based methods

In recent years, deep learning has gained enormous popularity in many fields, including computer vision [116], NLP [117] and speech recognition [118]. Its popularity is largely due to its capacity to capture features automatically, i.e., without the need for feature engineering. Deep learning has been applied to mental illness detection tasks, and has been shown to outperform conventional machine learning-based models. There are many different neural network structures, which include convolutional neural networks (CNN), recurrent neural networks (RNN), and transformer-based networks (especially pre-trained language representation models). Table 2 summarises the details of a number of deep learning-based models for mental illness detection. The main steps involved in the three types of fusion strategies mentioned in Table 2, i.e., feature-level fusion, model fusion and task fusion, are depicted in Fig. 5.

#### 3.2.1. Feature-level fusion

Similarly to conventional machine learning-based methods, many deep learning-based methods also begin by extract emotion features using NLP tools, after which the features are concatenated with textual embeddings or processed as new inputs for training neural network models. We term this type of fusion as *feature-level fusion* (also called early fusion) (see Fig. 5(a)).

There are many approaches to extracting emotion features by using existing emotion analysis tools. For instance, Fig. 6(a) illustrates the approach taken by Song et al. [119], who propose a feature attention network that leverages four different potential indicators (i.e., depressive symptoms, sentiments, ruminative thinking and writing style) to simulate the process of detecting depression and to increase the model's interpretability. For sentiments, the authors firstly obtain the sentiment categories (positive, neutral, and negative) for each word using SentiWordNet, and then encode the one hot vectors as sentiment features using an RNN. The other three features are also generated using basic neural networks and specific linguistics tools. Finally, an attention mechanism is also applied. Fig. 6(b) depicts the approach taken in Melvin et al. [120], which firstly calculates the subjective and sentiment scores of tweets using the TextBlob sentiment analysis library, and then concatenates these features with word vectors and a feature denoting the posting time to obtain the final text representations. They then used a Gated Recurrent Units (GRU) model to identify the sleep-deprivation status of users. In [121], the authors presented a cascade model named SNAP-BATNET (see Fig. 6(c)), which combines multiple handcrafted features, including author profiling, textual features, social network graph embeddings and tweet metadata. The emotion features are extracted using the NRC emotion lexicon, with ten scores (i.e, eight emotion categories and two polarities). A feature stacking based architecture is then used to combine these features from BiLSTM networks or Random Forest classifiers for suicide ideation detection. Fig. 6(d) represents the approach taken in [122,125]. Both of these studies extract and concatenate linguistic features and sentiment features obtained using NLP tools, which are then fed into CNN/LSTM structure for training. Uban et al. [19] explored the use of sophisticated deep learning architectures for the detection of mental disorders, which consist of a post-level encoder with CNN/LSTM, traditional emotion and BOW features that are passed to dense layers, and a user-level hierarchical attention network (see Fig. 6(e)). The model can fully leverage linguistic features at different levels, including both contextual information from texts, and emotions expressed by the user. Ablation experiments revealed both the importance of emotion features and the effectiveness of the hierarchical network structure. Fig. 6(f) illustrates the approach taken by Ren et al. [34] to designing an emotion-based attention network model. There are two parts in this model, i.e., a semantic understanding network (a standard BiLSTM module) used to obtain contextual semantic information, and emotion understanding network. For the emotion understanding network, the emotion encoding layer and BiLSTM-Attention layer firstly encode positive and negative words to obtain the corresponding positive and negative emotion information. Then, a dynamic fusion strategy with trainable floating parameter is applied to fuse emotion information. The experimental results revealed that emotion information is effective in detecting depression.

In addition to these six representative model structures, there are more publications on feature-level fusion. For example, Alhuzali et al. [36] used pre-trained language models to extract features. In particular, for emotion features, they leveraged SpanEMO model [133] that achieves competitive performance on the SemEval-2018 multi-label emotion classification dataset. Subsequently, a random forest classifier was trained on top of the many extracted features to predict depression severity. Ji et al. [126] proposed to fuse lexicon-based sentiment features into the hidden representation of recurrent neural networks by an attentive relation network for suicidal ideation and mental disorder classification. The authors adopted domain-specific sentiment lexicons and applied a relation network to the calculation of relational scores between sentiment and hidden features. The relational features, which are learned by the relation network and prioritised by an attention mechanism, act as the final fused features.

Other approaches leverage novel representations to obtain emotion-related features, including Bag of Sub-Emotions (BoSE) [134] and Deep Bag of Sub-Emotions (DeepBoSE) [124]. Fig. 7(a) provides a conceptual diagram of the BoSE representation. Firstly, based on the broad emotions obtained from a lexical resource [60] and sub-word embeddings obtained from FastText [135], a set of fine-grained emotions is generated by applying cluster algorithms. Then, texts are masked by replacing each word with the label of its closest fine-grained emotion and computing the frequency histogram of fine-grained emotion to obtain the BoSE representations. BoSE representations can better capture more specific emotions in documents than traditional BOW representations. Aragon et al. [123] used BoSE representation in conjunction with a deep attention model that includes CNN and GRU structures to detect depression and anorexia; experimental results demonstrated that the method outperforms basic deep learning models that do not make use of sub-emotion information. In addition, Lin et al. [35] also tried to combine BoSE representations with the recently proposed prompt-learning paradigm [136] for depression severity detection, which achieved competitive results. In order to overcome the disadvantages of BoSE (i.e., offline learning and separate training), the DeepBoSE representation [124] was proposed, as shown in Fig. 7(b). Lara et al. [124] presented a deep learning model that computes BoSE representations. In particular, a dissimilarity mixture autoencoder [137] and probabilistic differentiable Bag-of-Features were exploited to increase the interpretability of DeepBoSE representations, which also achieved better performance than the original BoSE for depression detection.

---

[7] https://apis.paralleldots.com/text_docs/index.html
[8] https://www.meaningcloud.com/

**Table 1**
Summary of machine learning-based approaches to mental illness detection that incorporate emotion fusion. *Abbreviations: SVM - Support Vector Machines, MLN - Markov Logic Networks, NB - Naive Bayes, LR - Logistic Regression, ANN - Artificial Neural Networks, RF - Random Forest, GP - Gaussian Process Classifier, HMM - Hierarchical Hidden Markov Model, GBDT - Gradient Boosted Decision Trees, FCM - Fuzzy Cognitive Map, IBPT - Inverse Boosting Pruning Trees, KNN - K-Nearest Neighbours, DT - Decision Tree, XGboost - eXtreme Gradient Boosting.

| Publication | Year | Source | Mental illness | Model | Emotion features extracted | Tools |
|---|---|---|---|---|---|---|
| [87] | 2012 | Forum | Mental disorders | SVM, MLN | Sentiment | LIWC |
| [88] | 2013 | Weibo | Depression | BayesNet, J48Tree, Decision Tree | Sentiment | HowNet [89] + some rules |
| [81] | 2013 | Twitter | Depression | SVM | Emotion | LIWC, ANEW |
| [82] | 2013 | Weibo | Mental health status | SVM | Emotion word ontology features | Chinese emotion word ontology dictionary |
| [90] | 2013 | Twitter | Depression | Probabilistic model | Emotion | LIWC, ANEW |
| [91] | 2014 | Weibo | Pressure | NB, SVM, ANN, RF, GP | Emotional degree | Stress-related linguistic lexicons |
| [92] | 2015 | Tumblr | Anorexia | SVM | Sentiment | SentiWordNet |
| [77] | 2015 | Tumblr | Suicide | SVM, J48Tree, NB, RF | Sentiment, emotion | LIWC, SentiWordNet |
| [83] | 2015 | Blog | Suicide | LR | Emotion | CET model [93] |
| [94] | 2016 | Twitter | Depression | SVM, ensemble model | Sentiment, mood of emoticons | SentiStrength, emoticon lexicons |
| [95] | 2016 | reachout | Mental disorders | SVM | Emotion | word2vec (Euclidean distance) |
| [86] | 2016 | Twitter | Suicide | Semi-supervised model | Emotion | hashtag + semi-supervised model |
| [96] | 2017 | Twitter | Depression | GBDT | Sentiment | SentiStrength |
| [97] | 2017 | Twitter | Stress | TensiStrength | Sentiment | LIWC, SentiStrength |
| [63] | 2017 | Reddit | Depression | SVM, KNN, RF, LR | Sentiment | VADER |
| [59] | 2018 | Livejournal | Depression | RF, HMM | Sentiment | LIWC, ANEW |
| [61] | 2018 | Twitter | Anxiety | LR | Sentiment, emotion | NRC Emotion Lexicon |
| [56] | 2018 | Facebook | Depression | SVM, KNN, DT, ensemble model | Emotion | LIWC |
| [78] | 2018 | Reddit | Suicide | SVM | Sentiment, emotion | LIWC, TextBlob |
| [98] | 2018 | Forum | Depression | LR | Sentiment | ANEW, LabMT [99] |
| [100] | 2018 | Facebook | Depression | rule-based model | Emotion variability and instability | LIWC + statistics |
| [101] | 2018 | Twitter | Depression | LR, SVM, NB, DT, RF | Emotion | LIWC, EMOTIVE |
| [66] | 2018 | Twitter | Mental disorders | SVM, RF, NB, DT | Emotion | EMOTIVE |
| [57] | 2018 | Reddit | Bipolar | SVM, LR, RF | Emotion | LIWC |
| [76] | 2018 | Reddit | Depression | RF | Sentiment, emotion | NRC Emotion Lexicon |
| [102] | 2019 | Weibo | Suicide | FCM | Emotion traits | CET model [83] |
| [103] | 2019 | Weibo | Depression | SVM | Emotion | Emotional dictionary and emoticon dictionary |
| [104] | 2019 | Twitter | Mental disorders | Sentiment polarities algorithm | Sentiment | Multipolarity sentiment affect intensity lexicon |
| [105] | 2019 | Twitter | Mental disorders | Rule-based Tree | Emotion | SenticNet, VADER, TextBlob |
| [80] | 2019 | Twitter | Suicide | RF, adaboost, BN, J48 | Sentiment | LIWC |
| [106] | 2019 | Twitter | Depression | IBPT | Sentiment | NLTK |
| [84] | 2020 | pantip | Depression | DT, RF, GBT | Emotion | ParallelDots |
| [107] | 2020 | Twitter | Depression | NB | Sentiment | VADER, TextBlob |
| [22] | 2020 | Twitter | Depression | SVM, LR, RD, GBDT, XGboost | Sentiment | LIWC, VADER |
| [79] | 2021 | Facebook | Depression | NB | Sentiment | Emotion dictionary |
| [71] | 2021 | Twitter | Suicide | Semi-supervised model | Sentiment, emotion | SentiStrength, NRC Affect Intensity Lexicon |
| [64] | 2021 | Twitter | Depression | NB, NBTree | Sentiment | TextBlob |
| [69] | 2021 | Twitter | Depression | LR | Sentiment | SentiWordNet |
| [108] | 2021 | Twitter | Psychological distress | SVM, LR, RF | Sentiment, emotion | LIWC |
| [109] | 2021 | Reddit | Self-harm | SVM | Sentiment | NLTK |
| [110] | 2021 | Reddit | Mental disorders | SVM, LR, RF | Emotion | BERT emotion classifier |
| [111] | 2021 | Reddit | Depression | SVM, LR | Sentiment | SentiWordNet |
| [85] | 2021 | Reddit | Mental disorders | adaboost, RF | Sentiment, emotion | VADER, MeaningCloud, ParallelDots |
| [74] | 2021 | Twitter | Depression | LR, SVM, DT, RF | Sentiment | SentiWordNet, SenticNet |
| [112] | 2022 | Reddit | Suicide | LR | Sentiment, emotion | Pre-trained BERT models (Bertweet-base-sentiment, EmoRoBERTa, Twitter-roberta-base-emotion) |
| [113] | 2022 | Reddit | Mental disorders | DT | Emotion | LIWC |
| [114] | 2022 | Twitter | Suicide | LR, SVM, RF, XGBoost | Sentiment | SentiWordNet |
| [115] | 2022 | Twitter | Suicide | RF | Sentiment | VADER |

### 3.2.2. Model fusion

In model fusion, the emotion information usually comes from the hidden features of other pre-trained emotion models. The overall model is an end-to-end framework without additional features extracted from NLP tools.

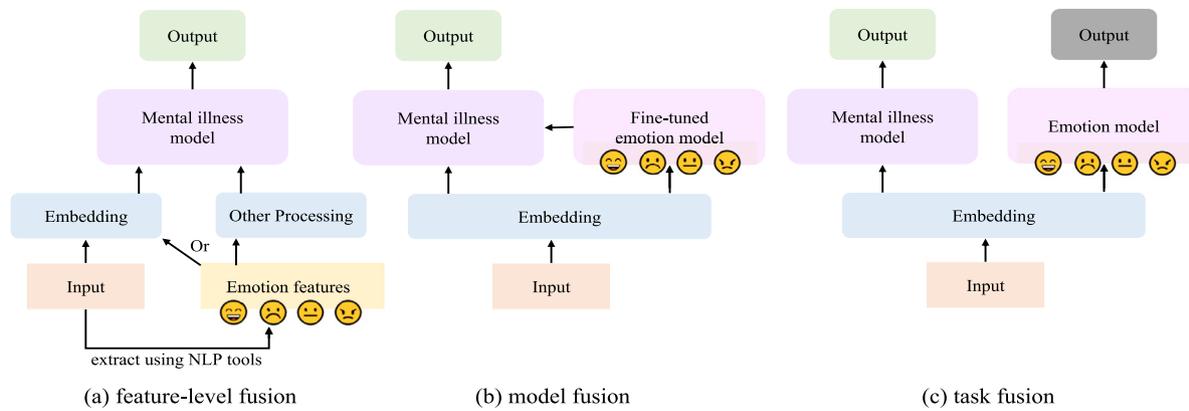

**Fig. 5.** Different emotion fusion strategies used in deep learning-based methods.

**Table 2**
Summary of deep learning-based methods with emotion fusion for mental illness detection.

| Publication | Year | Source | Mental illness | Fusion strategy | Method | Features |
|---|---|---|---|---|---|---|
| [119] | 2018 | Reddit | Depression | Feature-level fusion | Feature attention network | depressive symptoms, sentiments, ruminative thinking, writing style |
| [120] | 2019 | Twitter | Sleep deprivation | Feature-level fusion | GRU | Sentiment, subjectivity from TextBlob, post time |
| [121] | 2019 | Twitter | Suicide | Feature-level fusion | SNAP-BATNET | Emotion, sentiment |
| [122] | 2020 | Twitter | Psychological tendency | Feature-level fusion | CNN | Sentiment from SentiWordNet, some rules |
| [123] | 2020 | Reddit | Depression and anorexia | Feature-level fusion | Deep Emotion Attention Model | Sub-emotion embedding |
| [124] | 2021 | Reddit | Depression | Feature-level fusion | Deep Bag-of-Sub-Emotions | Sub-emotion embedding |
| [125] | 2021 | Twitter | Depression | Feature-level fusion | LSTM | Sentence-level emotion, VAD features |
| [19] | 2021 | Reddit | Mental disorders | feature-level fusion | CNN, BiLSTM-Attention | Emotion features from LIWC and NRC Emotion Lexicon |
| [34] | 2021 | Reddit | Depression | Feature-level fusion | BiLSTM-Attention | Emotion |
| [36] | 2021 | Reddit | Depression | Feature-level fusion | BERT | Emotion features from SpanEmo |
| [126] | 2022 | Reddit | Suicide | Feature-level fusion | Attentive relation networks | Sentiment |
| [35] | 2022 | Reddit | Depression | Feature-level fusion | Prompt-learning | Sub-emotion embedding |
| [127] | 2022 | reddit and twitter | Stress | Feature-level fusion | BiLSTM, BERT | Emotion and sentiment |
| [128] | 2022 | Reddit | Depression | Feature-level fusion | Gated multimodal networks | Emotion features from NRC Emotion Lexicon |
| [129] | 2022 | Reddit | Depression | Feature-level fusion | RoBERTa with contrastive learning | Sentiment from VADER |
| [28] | 2020 | Twitter | Suicide | Model fusion | STATENet | Plutchik Transformer fine-tune on Emonet |
| [29] | 2021 | Twitter | Suicide | Model fusion | Phase-aware emotion progression model | Plutchik Transformer fine-tune on Emonet |
| [30] | 2021 | Twitter | Suicide | Model fusion | Hyperbolic Graph Convolutional Network | Plutchik Transformer fine-tune on Emonet |
| [130] | 2022 | Twitter | Depression | Model fusion | ERAN | Emotion from pre-trained TextCNN |
| [131] | 2021 | Reddit | Stress | Model fusion/task fusion | Transfer learning | Emotion model fine-tune on Emonet |
| [132] | 2022 | Suicide notes | Depression | Task fusion | BiGRU-Attention multi-task learning | External sentiment and emotion knowledge |

Sawhney et al. [28–30] have undertaken a number of studies relating to the detection of suicidal ideation using the model fusion strategy. By taking into account historic information about emotions expressed in tweets in the dataset, they proposed a suicidality assessment time-aware temporal network (STATENet) [28], which is depicted in Fig. 8(a). The overall structure of STATENet consists of two main parts: individual tweet modelling and historic tweets modelling. They first used SentenceBERT [138], a pre-trained transformer model, to encode the tweet to be assessed. Meanwhile, to obtain a fine-grained emotional representation, they used a BERT model named PlutchikTransformer, which was trained on the Emonet dataset [139], consisting of approximately 1.6 million tweets labelled according to a set 24 fine-grained emotions. In addition, a Time-aware LSTM (T-LSTM) was designed, because time intervals could be useful for analysing changes in emotional states over time. It may be observed in Fig. 8(a) that in contrast to the traditional LSTM structure, T-LSTM introduces a time-lapsed parameter ($\Delta t$). Finally, STATENet concatenates the hidden semantic representations of the tweet to be assessed and the emotion representations from historic tweets, followed by the prediction layer.

In contrast to STATENet, the phase-aware suicidality identification emotion progression model (PHASE) [29] contains a new LSTM variant and also encodes the tweet to be assessed with PlutchikTransformer, as shown in Fig. 8(b). Inspired by the structure of stage-aware LSTM [140], Sawhney et al. designed a Time-Sensitive Emotion LSTM (TSE-LSTM), which introduces a new time-sensitive long-term gate and a short-term gate. Additionally, a cumulative sum (cumsum) operation was applied in the TSE-LSTM to take into account the influence of recent emotional context. Following the TSE-LSTM layer is a phase adaptive convolution layer, which can extract features from the learned phase-wise progression in the user's most recent emotional phase. In another study [30], they proposed Hawkes temporal emotion aggregation mechanism with Hawkes [141] Process(HEAT), whose architecture is shown in Fig. 8(c). This architecture is based on the assumption that emotions expressed in different historical tweets can influence each other. Moreover, a heterogeneous social network graph and hyperbolic graph convolution network are employed.

### 3.2.3. Task fusion

The goal of task fusion is to leverage a multi-task framework to enable the model to learn the features for auxiliary tasks (e.g., emotion classification or sentiment classification), thus improving the performance of the primary task (i.e., mental illness detection).

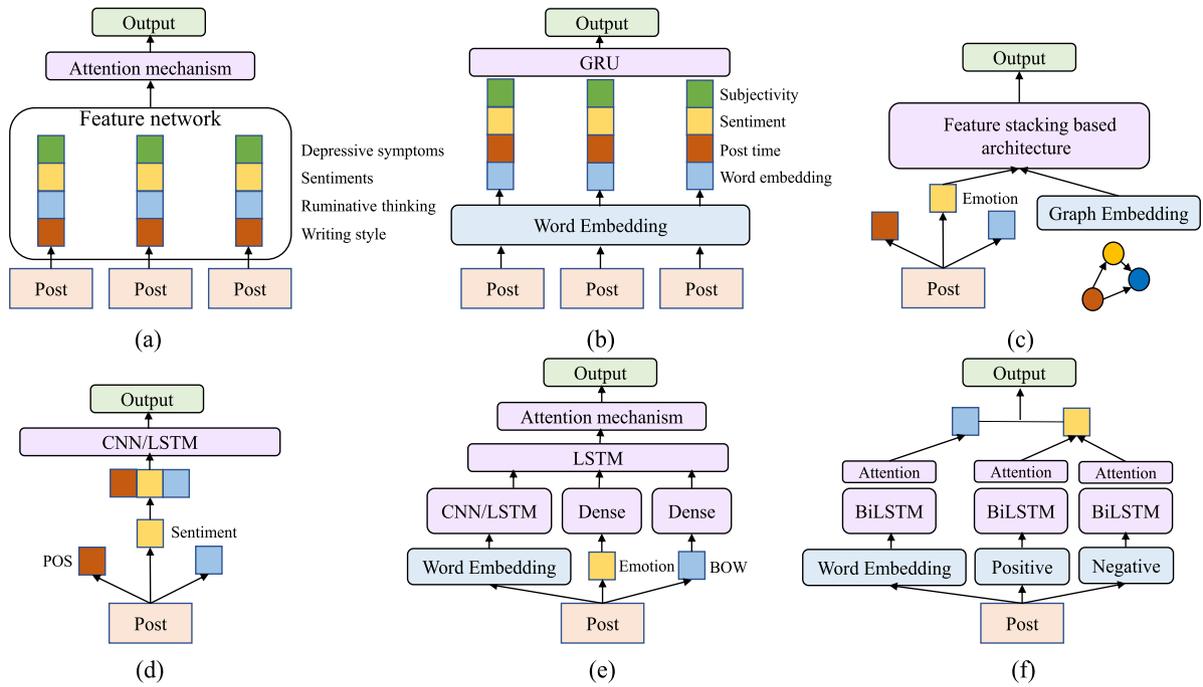

Fig. 6. Different model structures based on feature-level fusion. (a) FAN [119], (b) Melvin et al. [120], (c) SNAP-BATNET [121], (d) [122,125], (e) HAN [19], (f) SUN-EUN [34].

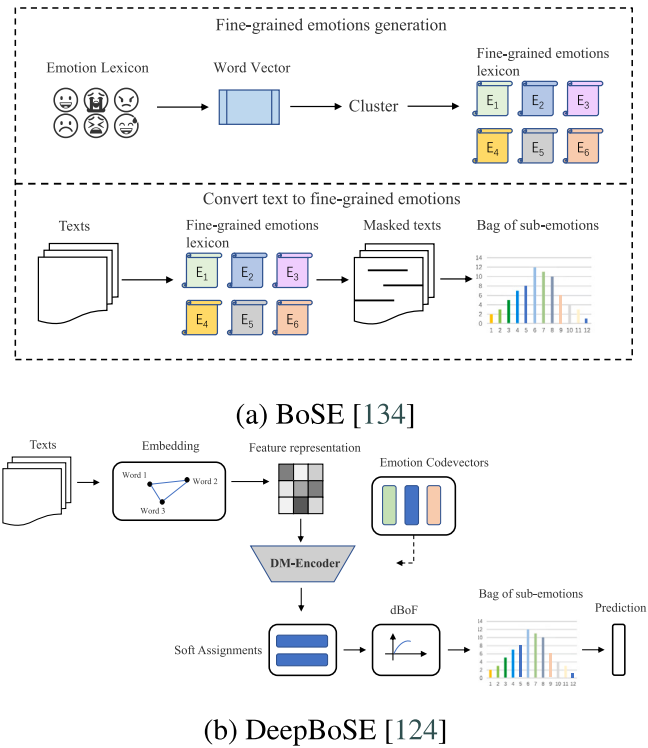

Fig. 7. Conceptual diagram of the BoSE representation and the DeepBoSE architecture.

Turcan et al. [131] explored the application of emotion-infused multi-task learning models to the task of stress detection; the different architectures employed are illustrated in Fig. 9. Fig. 9(a) depicts an alternating multi-task model, which is a framework in which two single-task models share the same BERT layers when training. The stress task was trained with Dreaddit dataset [142] and the emotion task was trained with GoEmotions [143]. In contrast, in the classical multi-task model (Fig. 9(b)), two tasks are trained at the same time with the same dataset. The authors firstly used a fine-tuned emotion BERT model to predict the emotion labels for each stress tweet, and trained on new BERT model with these emotion labels and the ground truth stress label of Dreaddit. However, the fine-tuned model (Fig. 9(c)) utilises a model fusion strategy for stress prediction by leveraging the fine-tuned emotion model.

To study the effects of different multi-task learning models, Ghosh et al. [132] designed a number of multi-task architectures. Fig. 10(a) is the basic multi-task architecture. It can be observed in Fig. 10(b) that the task-specific learned features from the depression and sentiment tasks are concatenated with original features encoded by BiGRU, and the multi-feature fusion vector is fed into a dense layer to predict the final emotion. The knowledge-infused multi-task architecture (Fig. 10(c)) contains an external knowledge module in which SenticNet's IsaCore and AffectiveSpace tools are exploited to introduce commonsense knowledge. Then, the authors concatenated the two external features with the original features and fed the vector into the cascaded multi-task architecture. Although this paper considers emotion recognition as the primary task, depression detection and sentiment classification as auxiliary tasks, which is different from general emotion-based task fusion, the experimental results also prove that the multi-task models perform better than the single-task models. In addition, the performance of the cascaded architecture and knowledge-fused architectures is better than that of the basic multi-task architecture.

## 4. Discussion

We have reviewed a large number different mental illness detection models and emotion fusion strategies. In this section, we discuss the following aspects of these models and strategies:

(1) Effects of feature engineering-based methods and deep learning-based methods;
(2) Effects of emotion fusion;
(3) Effects of different fusion strategies.

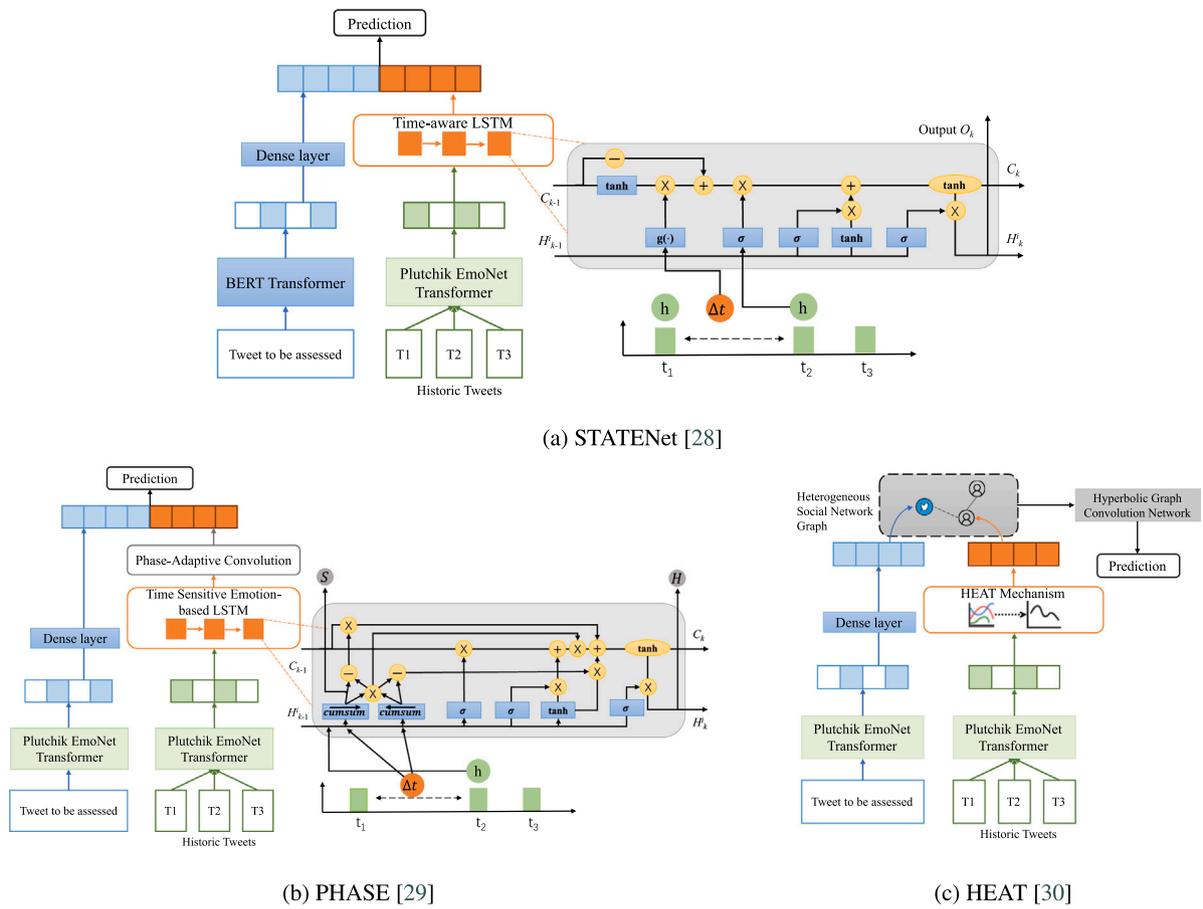

Fig. 8. Examples of end-to-end model fusion architectures.

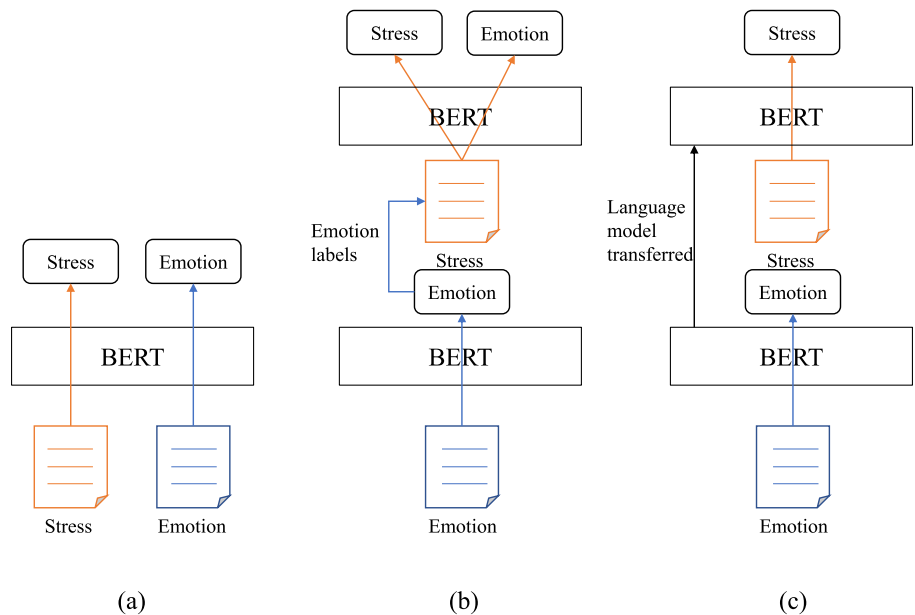

Fig. 9. Emotion-informed architectures for stress detection. (a) is an alternating multi-task model; (b) is a classical multi-task model with hard parameter sharing; (c) is a fine-tuned model.

## 4.1. Effects of feature engineering-based methods and deep learning-based methods

As explained in the previous section, most of the early approaches to mental illness detection using emotion fusion were based on feature engineering models. The feature engineering-based method combines the extraction of hand-crafted features with well-designed machine learning models. The extraction of these features relies on the application of different NLP tools. For example, dependency parsing may be used to extract syntactic features, while SentiWordNet may be used to extract emotion features [77]. It is inevitable that the quality of the extracted features will depend on the specific NLP tools that are employed; this in turn will affect the results of the final classification to some extent. SVM classifiers are commonly used in feature engineering based methods because they can achieve relatively high performance on the majority of datasets.

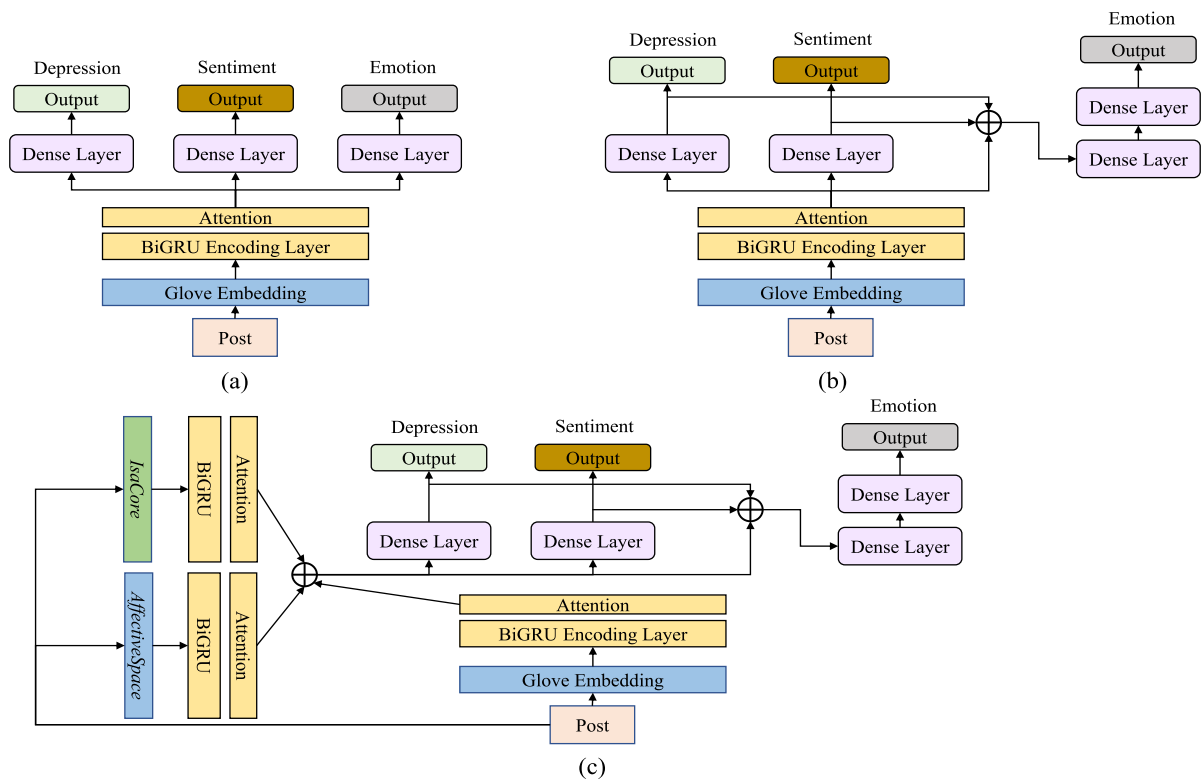

**Fig. 10.** Architectures of different multi-task learning models. (a) is a basic multi-task architecture with hard parameter sharing; (b) is a cascaded multi-task architecture; (c) is a knowledge-infused multi-task architecture.

Recently, deep learning-based methods have grown in popularity and have been applied to many NLP tasks [144] due to their strong feature learning ability, which usually outperforms conventional machine learning-based methods. Compared to feature engineering-based methods, deep learning-based methods do not rely heavily on feature engineering, but instead use neural networks to automatically capture valuable features. It can be observed from Table 2 that pre-trained language models like BERT [145] and MentalBERT [12] have received large amounts of attention in the last two years, according to their ease of use and strong linguistic representations, which result in high levels of performance on mental illness detection tasks.

*4.2. Effects of emotion fusion*

Social media texts often convey emotions whose detection can improve the performance of mental illness surveillance. In order to explore the effectiveness of emotion fusion, experimental results achieved by various deep learning-based methods that incorporate emotion fusion are summarised in Table 3. The table includes the results of emotion ablation studies (where available), which illustrate that incorporating emotion features into the models can enhance their performance [28,34,36]. The ablation study of [125] demonstrates that the effects of emotion fusion are stronger than those of topic fusion and fusion of users' online behaviour; this provides strong evidence that emotion is one of the most important factors in the judgment of mental illness. Additionally, the use of particular fusion network designs (like T-LSTM [28] and TSE-LSTM [29]) can also lead to improved results.

*4.3. Effects of different fusion strategies*

As introduced above, there are three basic emotion fusion strategies for mental illness detection, i.e., feature-level fusion, model fusion and task fusion.

The results of our survey reveal that feature-level fusion is the mostly employed strategy, in which a concatenation operation is usually used to combine the semantic features with emotion features obtained from NLP tools. The process is easy to implement and can exploit the interaction and correlation between different low-level features. Despite the simplicity of this strategy, it is still necessary to use external tools to extract emotion features. In addition, directly appending emotion features to word-embedding features may not be efficient because their vector spaces are distinct, which will generate redundant input vectors [152].

For model fusion, the introduction of fine-tuned emotion models fuses hidden emotion features, which significantly improves performance than feature-level fusion [28,29]. On the other hand, the final results are affected by the accuracy of the emotion classification model that has been introduced [153]. If the fine-tuned emotion model does not perform well, it will generate noise or erroneous hidden features, which will have a negative impact on the performance of detecting mental illness. Hence, it is important to choose an appropriate model; if possible, it is desirable to use the original data (having both emotion labels and mental illness labels) for fine-tuning, since this can reduce the impact of different training data [131].

Task fusion is achieved using a multi-task learning framework in which parameters of certain layers are shared between multiple tasks. Task fusion can reduce overfitting and help to improve learning of auxiliary information [154]. A major challenge of applying task fusion to mental illness detection, however, is the lack of suitable multi-task datasets. Currently, there are only two datasets that focus on mental illness and the role of emotions, i.e., the CEASE corpus, which is concerned with suicidal ideation [151] and an emotion annotated mental health dataset (EmoMent), whose data originates from two South Asian countries [155].

Table 3

Performance summary of deep learning-based methods with emotion fusion for mental illness detection.

| Publication | Dataset | Performance | Emotion ablation study |
|---|---|---|---|
| [119] | RSDD [146] (9,210 depression, 107,274 control) | F1 0.56<br>Baseline BiRNN 0.54 | |
| [120] | UCLA student tweets (8,068 sleep deprived tweets, 10,326 control) | AUC 0.68 | |
| [121] | Private dataset collected from Twitter (3,984 suicidal tweets, 30,376 control) | F1 0.923 AP 0.709<br>Baseline RCNN 0.921 0.704 | Use NRC features 0.891 0.641<br>no NRC features 0.891 0.640 |
| [122] | Private dataset collected from Twitter (2,880 positive, 1,883 neutral, 2,605 negative) | F1 0.76<br>Baseline lexicon-based method 0.71 | |
| [123] | eRisk 2018 task (214 depression, 1,493 control)<br>eRisk 2019 task (134 anorexia, 1,153 control) | F1 (depression) 0.58 (anorexia) 0.79<br>Baseline DeepAttention 0.50 0.66 | |
| [124] | eRisk 2017 task (135 depression, 752 control)<br>eRisk 2018 task (214 depression, 1,493 control) | F1 0.6415 0.6545<br>Baseline BoSE 0.6359 0.6316 | |
| [125] | Depression dataset curated by [147] (1,402 depressed, 5,160 control) | MSE 1.43<br>Baseline SVM 1.76 | Only use emotion features 1.50<br>only use topic features 1.68<br>only use online behaviour 1.80 |
| [19] | eRisk task (214 depression, 1,090 control)<br>eRisk 2019 task (134 anorexia, 1,153 control)<br>eRisk 2020 task (145 self-harm, 618 control) | F1 (depression) 0.65 (anorexia) 0.61 (self-harm) 0.45<br>Baseline BiLSTM 0.62 0.53 0.40 | Use emotion features 0.65 0.61 0.45<br>no emotion features 0.59 0.45 0.43 |
| [34] | Reddit data [148] (1,293 depression posts, 5,49 control) | F1 0.94<br>Baseline BiLSTM+attention 0.926 | Use emotion features 0.94<br>no emotion features 0.925 |
| [36] | eRisk 2021 depression task (minimal 20, mild 40, moderate 49, severe 61) | Depression Category Hit Rate (DCHR) 22.5 | Use SpanEmo features 22.5<br>no SpanEmo features 21.25 |
| [126] | UMD dataset [149] (363 low, 503 severe, 866 control)<br>private SWMH dataset (different mental disorders 54,412) | F1 0.545 0.648<br>Baseline BiLSTM 0.523 0.619 | Use sentiment features 0.824<br>no sentiment features 0.820 |
| [35] | DepSign shared task (not 3,081, moderate 8,325, severe 1261) | F1 0.582<br>Baseline RoBERTa 0.512 | |
| [28] | #suicidal dataset [150] (3,984 suicidal tweets, 30,322 control) | F1 0.799<br>Baseline DualContextBERT 0.767 | Use Plutchik emotion BERT 0.778<br>no Plutchik emotion BERT 0.767 |
| [29] | The same as above | F1 0.805<br>Baseline DualContextBERT 0.767 | Use TSE-LSTM 0.796<br>use tradition LSTM 0.780 |
| [30] | The same as above | F1 0.792 | |
| [131] | Dreaddit [142] (3,553), Goemotion (58,009 27 emoition) | F1 alterating multi-task models 0.802<br>classical multi-task models 0.803<br>fine-tuning model 0.803<br>BERT 0.789 | |
| [132] | CEASE [151] (4,932) | Accuracy 0.744<br>Baseline single-task 0.753 | |
| [127] | Dreaddit | F1 0.860<br>Baseline BERT 0.848 | |
| [128] | eRisk 2018 task (214 depression, 1,493 control) | F1 0.70<br>Baseline SVM 0.63 | |
| [129] | DepSign shared task (not 3,081, moderate 8,325, severe 1261) | Macro-F1 0.552<br>Baseline pre-trained models 0.528 | |
| [130] | TTDD dataset [147] (1,402 depression, 1,402 control) | F1 0.904<br>Baseline BERT 0.853 | use emotion features 0.904<br>no emotion features 0.873 |

## 5. Challenges and future directions

Although this review has amply demonstrated the power of emotion fusion in mental illness detection, several challenges still remain. In this section, we present some of the most important challenges and outline potential future research directions for detecting mental illness.

### 5.1. Availability and quality of datasets

Most of the methods that this review has identified are based on supervised learning frameworks, which require annotated training data. Accordingly, the availability of suitable high quality annotated datasets is critical to the success of these methods. However, the sensitive nature of mental illness, combined with strict ethical protocols, has a large impact on the availability of data. Table 4 provides an overview of mental illness detection databases from emotion fusion methods over the past five years, including the commonly used name and reference, the data source, types of mental illness, the number of samples, the availability and the annotation quality. We find that the majority of datasets are private, and only some are publicly available or from shared tasks. Meanwhile, many datasets are not manually annotated by professional psychiatrists, but rather weakly labelled by using some rules, such as subreddit names and self-declared statement expressions (e.g., "I was/have been diagnosed with depression") [18]. Therefore, particular issues include the following: (1) There are not enough public datasets available, which limits the number and scope of studies relating to mental illness detection; (2) the size and quality of the datasets are variable; (3) some issues such as annotation bias can lead to difficulty in training reliable models [4]. To alleviate the above problems, we encourage the researchers to share their datasets and code, while taking into account ethical considerations to ensure privacy protection and to avoid psychological distress [156,157]. It also can be seen that, among the 38 datasets from Table 4, depression (42%) and suicide (24%) constitute the majority of types of mental illnesses, followed by other mental disorders (34%). Hence, establishing different mental illness detection datasets (e.g., anorexia and bipolar) should be encouraged, meeting future demands for mental health applications. It is also be desirable to develop further mental illness detection datasets which include emotion labels, in order to better support the evaluation and analysis of emotion fusion algorithms. Moreover, semi-supervised [158, 159] and weakly supervised methods can also be applied when the size of annotated datasets is insufficient to support fully supervised learning.

**Table 4**
Summary of mental illness detection datasets from emotion fusion methods over the past five years. *Abbreviations: A - Available, ASA - Available via Signed Agreement, ST - Shared Task dataset that need to be requested, P - Private dataset that need to be requested.

| Dataset | Source | Mental illness | Number of samples | Availability | Annotation quality |
|---|---|---|---|---|---|
| [59] | Livejournal | Depression | 2,019 depression, 2,007 control | P | Annotated according to subreddit name |
| [56] | Facebook | Depression | 4,149 depression, 2,996 control | P | No mention |
| [78] | Reddit | Suicide | 205 suicidal users, 580 control | P | Only a small part were manually annotated by psychiatrists |
| [98] | Forum | Depression | 749 participants | P | Survey on a crowdsourcing platform |
| [100] | Facebook | Depression | 78 users | P | Collected from mood-tracking App |
| [101] | Twitter | Depression | 585 depression, 600 control | P | Annotated according to self-declared statement expression |
| [66] | Twitter | Mental disorders | 1,372 different mental disorders, 6,596 control | P | Annotated according to self-declared statement expression |
| [57] | Twitter | Bipolar | 3,488 bipolar, 3,931 control | P | Annotated according to self-declared statement expression |
| [102] | Weibo | Suicide | 65 suicidal users, 65 control | P | Annotated by 3 psychological college students |
| [103] | Weibo | Depression | 135 depression, 252 control | P | Annotated by psychiatrists |
| [104] | Twitter | Mental disorders | 396 users who have mental disorders, 400 control | P | Annotated according to self-declared statement expression |
| [105] | Twitter | Mental disorders | 489 users | P | Annotated according to self-declared statement expression |
| [80] | Twitter | Suicide | 115 suicidal users, 172 control | P | From popular persons that committed suicide worldwide |
| [79] | Facebook | Depression | 7,146 comments | P | Annotated by calculating negative words scores |
| [71] | Twitter | Suicide | 4,987 tweets | P | Collected according to subreddit name and then identified by 2 experts |
| [69] | Twitter | Depression | 309 participants | P | Survey completed by participants |
| [108] | Twitter | Psychological distress | 639 psychological distress, 649 control | P | Annotated according to self-declared statement expression |
| [110] | Reddit | Mental disorders | 1,997 users | P | Annotated according to self-declared statement expression |
| [113] | Reddit | Mental disorders | 18,287 addiction, 11,385 anxiety, 16,442 depression | P | Annotated according to subreddit name |
| [114] | Twitter | Suicide | 445 suicidal users, 724 control | P | No mention |
| [148] | Reddit | Depression | 1,293 depression, 5,49 control | P | Annotated according to self-declared statement expression |
| [120] | Twitter | Sleep deprivation | 8,068 sleep deprived tweets, 10,326 control | P | Survey completed by participants |
| [121] | Twitter | Suicide | 3,984 suicidal tweets, 30,376 control | P | Annotated by 2 students in clinical psychology |
| [122] | Twitter | Psychological tendency | 2,880 positive, 1,883 neutral, 2,605 negative | P | Annotated by 3 annotators |
| eRisk 2018 task [160] | Reddit | Depression | 214 depression, 1,493 control | ST | Annotated according to self-declared statement expression |
| eRisk 2019 task [161] | Reddit | Anorexia | 134 anorexia, 1,153 control | ST | Annotated according to self-declared statement expression |
| eRisk 2020 task [162] | Reddit | Self-harm | 145 self-harm, 618 control | ST | Annotated according to self-declared statement expression |
| eRisk 2021 task [163] | Reddit | Depression | 20 minimal, 40 mild, 49 moderate, 61 severe | ST | Survey completed by participants |
| DepSign shared task [164] | Reddit | Depression | 8,325 moderate, 1,261 severe, 3,081 control | ST | Annotated by 2 domain experts |
| UMD [149] | Reddit | Suicide | 363 low, 503 severe, 866 control | P | Crowdsourced annotation and random selection for expert identification |
| SWMH [126] | Reddit | Mental disorders | 54,412 posts with different mental disorders | ASA | Annotated according to subreddit name |
| C-SSRS [165] | Reddit | Suicide | 500 users with different suicide risk severity | A | Annotated by 4 clinical psychiatrists |
| Eye's dataset[a] | Twitter | Depression | 2,314 depression, 8,000 control | A | Annotated according to word 'depression' |
| TTDD [147] | Twitter | Depression | 1,402 depression, 300 million control | A | Annotated according to self-declared statement expression |
| RSDD [146] | Reddit | Depression | 9,210 depression, 107,274 control | ASA | Annotated by 3 layperson annotators |
| #suicidal [150] | Twitter | Suicide | 3,984 suicidal tweets, 30,322 control | P | Annotated by 2 students in clinical psychology |
| Dreaddit [142] | Reddit | Stress | 3,553 posts | A | Crowdsourced annotation |
| CEASE [151] | Suicide notes | Depression | 4,942 notes | ASA | Annotated by 3 annotators |

[a] https://www.kaggle.com/datasets/bababullseye/depression-analysis

*5.2. Algorithm performance*

While our survey has shown that emotion fusion improves the performance of mental disorder prediction, certain models could still suffer from instability. Possible reasons include poor performance of the fine-tuned emotion model itself, and/or different training data sources (which may include diverse writing styles and topics). In order to develop more effective of deep learning and fusion strategies, it is important to design novel fusion structures or neural networks that can better fuse mutual information. Moreover, emotional or psychological theories could be applied to better capture and utilise emotion features.

An example is emotion dynamics [166], which quantifies the changes in a user's emotional states that occur over time. Furthermore, the nature of social media data posts means that, in addition to texts, it can also be important to take into account images and emojis that are included in posts, since these can also reflect the user's emotions [167,168]. Multimodal sentiment analysis is a growing field, for which new fusion approaches are required [169–171].

*5.3. Interpretability*

The purpose of detection models is to facilitate an understanding of a user's mental state and to predict mental illness in a timely manner. A good model should not only exhibit high levels of performance, but should also possess strong interpretability, since this can be valuable in guiding psychiatrists to better understand how certain textual features can be indicative of mental illness [25]. Although some complex emotion fusion models have achieved high levels of performance, they fail to explain the role of information pertaining to emotions in determining their predictions. In future research, it is recommended that collaboration with mental health professionals should firstly be strengthened through clinical self-report emotional examination (e.g., using interviews and questionnaires) combined with monitoring of emotions using social media posts. Moreover, we could leverage other resources or tools (such as knowledge graph [172,173] and emotion-cause pair extraction [174]), to help to explain the causes or triggers of different emotions.

## 6. Conclusion

The use of social media for mental illness detection has received increasing attention over recent years. In this article, we have provided a comprehensive survey of how emotion fusion has been used to support this task. After having identified relevant studies, we firstly classified the fusion strategies employed into feature engineering-based methods and deep learning-based methods. For the former class of methods, we introduced different conventional machine learning methods and various emotion feature extraction tools that are used to facilitate emotion fusion. Meanwhile, in terms of deep learning-based methods, we divided emotion fusion approaches into feature-level fusion, model fusion and task fusion. We summarised the structures and characteristics of these different fusion strategies in detail, and discussed both their advantages and disadvantages. In general, the results of our survey demonstrate that emotion fusion is a highly effective technique. To conclude the survey, we outlined the challenges of emotion fusion and also provided suggested directions for future research direction. It is intended that our study will help researchers to better apply and understand emotion fusion in the context of mental illness detection.

**CRediT authorship contribution statement**

**Tianlin Zhang:** Writing – original draft, Conceptualization, Methodology, Formal analysis, Data curation, Visualization, Writing – review & editing. **Kailai Yang:** Writing – original draft, Writing – review & editing. **Shaoxiong Ji:** Writing – review & editing. **Sophia Ananiadou:** Writing –review & editing, Funding acquisition.

**Declaration of competing interest**

The authors declare that they have no known competing financial interests or personal relationships that could have appeared to influence the work reported in this paper.

**Data availability**

No data was used for the research described in the article.

**Acknowledgements**

We would like to thank Paul Thompson for his valuable comments. This work is supported in part by funds from MRC, United Kingdom MR/R022461/1 and the Alan Turing Institute, United Kingdom and the National Centre for Text Mining, United Kingdom.

*T. Zhang et al.*                                                                                             *Information Fusion 92 (2023) 231–246*

...